\documentclass{vgtc}                       

\graphicspath{{figures/}{pictures/}{images/}{./}} 

\usepackage{times}                     

\usepackage{tabu}                      
\usepackage{booktabs}                  
\usepackage{lipsum}                    
\usepackage{mwe}                       

\usepackage{mathptmx}  

\usepackage{amsmath}
\usepackage{cuted}
\usepackage{enumitem}

\author{Ivan Snegirev$^{\S}$\thanks{e-mail: Ivan.Snegirev@skoltech.ru}\hspace{2mm}$^{\star}$ %
\and Elizaveta Semenyakina$^{\S}$\thanks{e-mail: Elizaveta.Semenyakina@skoltech.ru}\hspace{2mm}$^{\star}$
\and Mikhail Konenkov$^{\S}$$^{\ddagger}$\thanks{e-mail:Mikhail.Konenkov@skoltech.ru}\hspace{2mm}$^{\star}$
\and Artem Lykov$^{\ddagger}$$^{\S}$\thanks{e-mail: Artem.Lykov@skoltech.ru}
\and Miguel Altamirano Cabrera$^{\S}$$^{\ddagger}$\thanks{e-mail: M.Altamirano@skoltech.ru}
\and Dzmitry Tsetserukou$^{\S}$\thanks{e-mail: D.Tsetserukou@skoltech.ru}}
\affiliation{\scriptsize $^{\S}$Intelligent Space Robotics Lab, Skolkovo Institute of Science and Technology, Moscow, Russian Federation \\ $^{\ddagger}$R\&D Center, MWS, Moscow, Russian Federation \\ $^{\star}$These authors contributed equally to this work}

\title{ORCESTRA: VLM-driven Visual Robot programming in Mixed Reality}

\abstract{%
ORCESTRA is a mixed-reality system for programming robot digital twins through no-code waypoint teaching and language-guided control. In a passthrough mixed-reality workspace, users place robot twins on real surfaces, teach trajectories, save robot-relative episodes, or issue spoken/typed commands that a vision-language model converts into structured digital-twin plans. Both interaction modes share a backend for metric grounding, embodiment-aware validation, preview, confirmation, and digital-twin execution. The system supports heterogeneous robot embodiments, including fixed-base manipulators, a mobile base, and a humanoid robot, demonstrating MR validation as a safety layer for language-guided robot programming before physical deployment.%
}


\teaser{
  \centering
  \includegraphics[width=0.9\linewidth]{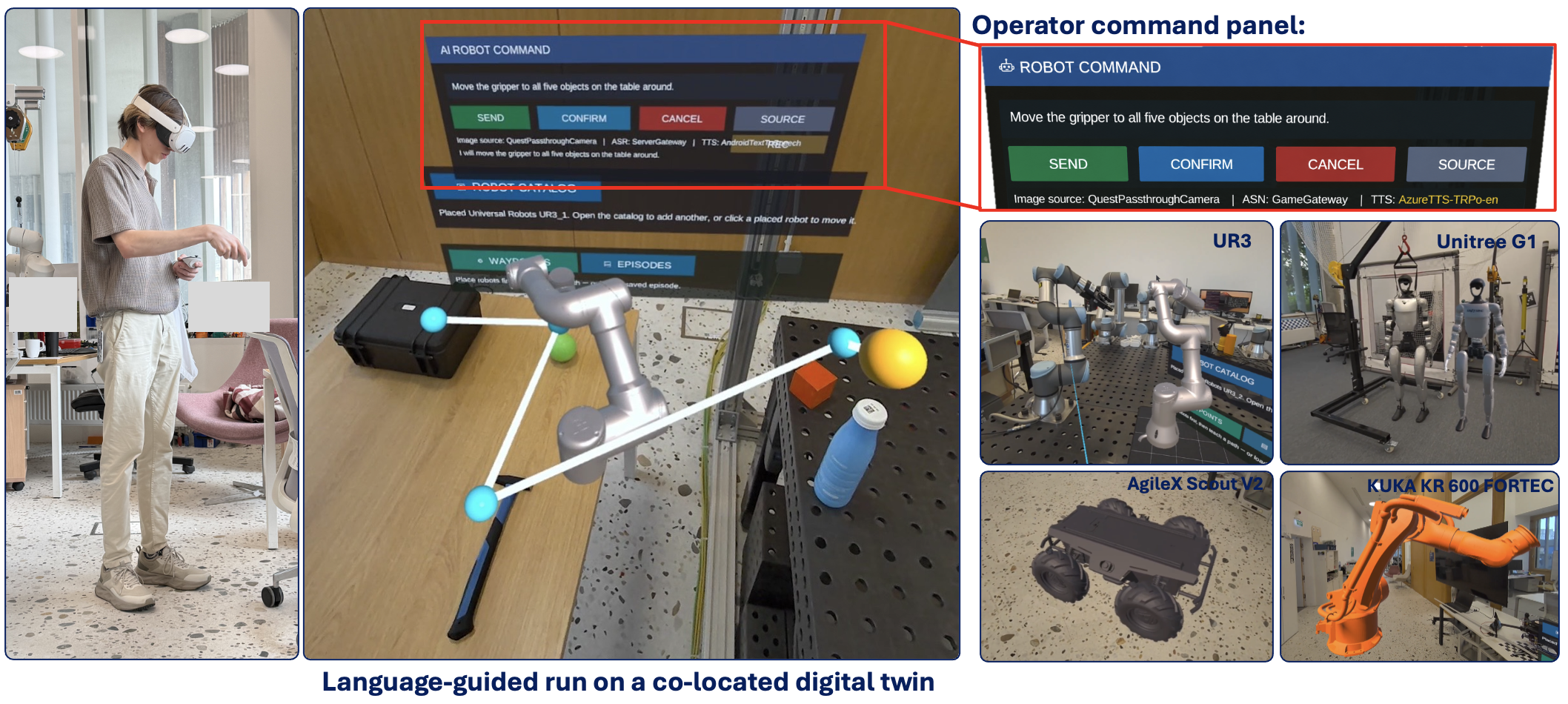}
  
  \caption{ORCESTRA in use. Left: a language-guided run on a co-located digital twin – the operator's command (top inset) is grounded by the VLM as waypoints along an approach spline, executed only after the operator confirms or manually refines the preview. Right: the four supported robot embodiments – UR3, KUKA KR~600~FORTEC, AgileX Scout~V2, and Unitree~G1.}
  \label{fig:overview}
}

\begin{document}

\firstsection{Introduction}
\maketitle

Programming robots directly on physical hardware remains costly and risky: small errors can cause unintended motion, collisions, or robot damage, and each ``specify -- observe -- correct'' iteration may require downtime on real equipment. Mixed-reality (MR) manipulator-programming interfaces~\cite{ostanin2020mr} and VR-based digital-twin frameworks~\cite{mr_cobot_dt} have therefore explored shifting robot programming away from physical hardware.

A co-located MR digital twin offers a useful intermediate representation as an interactive programming and verification workspace. Spatially aligned with the real cell through passthrough, it lets the operator inspect planned motion against the real environment – the table, equipment, walls, and nearby objects – and check, before execution, that the trajectory fits the available space and stays collision-free. Behaviors can thus be authored, tested, and corrected on the twin while the physical robot remains available, reducing downtime and dependence on vendor-specific teach-pendant interfaces.

This verification layer matters even more when behavior is specified through natural language using a Vision--Language Model (VLM). A VLM allows high-level commands such as ``move the object from one table to the other,'' but its output is less deterministic than manual teaching: the model may localize the wrong object, infer an incorrect destination, or choose an unreachable pose. A
co-located twin can ground, preview, edit, and confirm such plans before execution – especially important for humanoid robots, where an incorrect target may affect whole-body posture, self-collision, and human proximity, as also motivated by  HumanoidVLM~\cite{humanoidvlm}.

We present ORCESTRA, an MR system that unifies no-code waypoint teaching and VLM-guided task specification in a single co-located digital-twin environment, running standalone on a Quest~3 headset. In the language-guided mode the VLM converts the command into a structured plan, and the client renders it as an editable preview the operator refines before confirming. Both modalities share one backend for grounding, validation, preview, confirmation, and execution.

ORCESTRA is well suited to flexible manufacturing and small-batch production, laboratory and warehouse automation, and assistive or service robotics, where robots must be reprogrammed frequently and safely by non-expert operators. Its no-code and language-guided interfaces also make it a natural fit for robotics education and operator training, where learners author and verify behaviors on co-located twins without risk to hardware or bystanders. More broadly, the same MR workflow provides a low-cost prototyping and verification environment for research across heterogeneous \textbf{}

The key contributions of this work are:
\begin{enumerate}[
  label=\arabic*),
  leftmargin=3.2em,
  labelwidth=1.6em,
  labelsep=0.5em,
  itemsep=3pt,
  topsep=3pt,
  parsep=0pt,
  align=left
]
  \item \emph{A unified MR robot-programming environment} for heterogeneous digital twins – fixed-base manipulators, a mobile base, and a humanoid – registered through a shared abstraction.

  \item \emph{A dual-modality task interface} combining no-code waypoint teaching with VLM-guided language control in one co-located workspace.

  \item \emph{A shared typed-plan backend} for grounding, validation, editable preview, and confirmation-gated execution of manual and language-generated plans.

  \item \emph{A portable robot-relative episode format} for saving, replaying, repositioning, correcting, and future export of validated behaviors.
\end{enumerate}

\section{Related Work}

\subsection{Immersive Robot Programming and Digital Twins}

Simulation and digital twins let robot behaviors be designed, tested, and debugged before execution on hardware, reducing the cost and risk of trial-and-error on real robots. VR-based digital-twin frameworks for industrial robot programming~\cite{mr_cobot_dt} and warehouse-scale supervision systems such as WareVR~\cite{warevr} show how immersive digital twins support behavior authoring, monitoring, and correction; surveys of AR and robotics chart this design space more broadly~\cite{suzuki2022arsurvey}.

However, many VR digital-twin systems remain separated from the operator's physical workspace, observing the robot and environment in a purely virtual scene. Augmented- and mixed-reality interfaces instead overlay the robot onto the real workspace: head-mounted AR has been used to author robot trajectories in situ with waypoints and motion previews~\cite{quintero2018augtraj}, and
mixed-reality manipulator programming through a spatially registered digital twin~\cite{ostanin2020mr} lets the user inspect motion against real surfaces and nearby objects. ORCESTRA builds on this MR direction with passthrough co-location of heterogeneous embodiments, combining no-code teaching with VLM-guided task
specification.

\subsection{Language and Vision-Guided Robot Control}

Vision--language models can ground free-form instructions in visual scenes and produce structured task descriptions; in ORCESTRA this role is implemented with Qwen3-VL~\cite{qwen3vl}. ReKep~\cite{rekep} represents manipulation tasks as
relational keypoint constraints, while VoxPoser~\cite{voxposer} expresses robot objectives as composable 3D value maps, and vision--language guidance has also been applied to contact-rich humanoid manipulation~\cite{humanoidvlm}. Closest to our setting, recent work combines an LLM with mixed reality to
generate robot waypoints and preview them in AR~\cite{fang2024waypoint}. Such approaches make task specification more natural but introduce behavioral uncertainty; ORCESTRA differs by treating the model output not as a direct waypoint source but as a typed proposal to be grounded, embodiment-validated, previewed, and confirmed on the MR client, as detailed in Section~\ref{sec:system}.

\section{System Overview}
\label{sec:system}

ORCESTRA is implemented as a standalone Unity~6 application (Universal Render Pipeline, OpenXR). It follows a client--gateway architecture: the Quest~3 MR client maintains the co-located digital twins, interaction logic, scene grounding, preview, validation, and execution, while a lightweight FastAPI
gateway connects it to a VLM served through an OpenAI-compatible endpoint (Qwen3-VL via vLLM/SGLang in our deployment). The system spans five components – an MR digital-twin workspace, an embodiment-specific control layer, a no-code teaching interface, a VLM-guided specification pipeline, and a shared grounding, validation, preview, and execution layer – and takes either a taught motion or
a natural-language command through this shared path, with the twin moving only after explicit confirmation.

\begin{figure*}[t]
  \centering
  \includegraphics[width=0.75\textwidth]{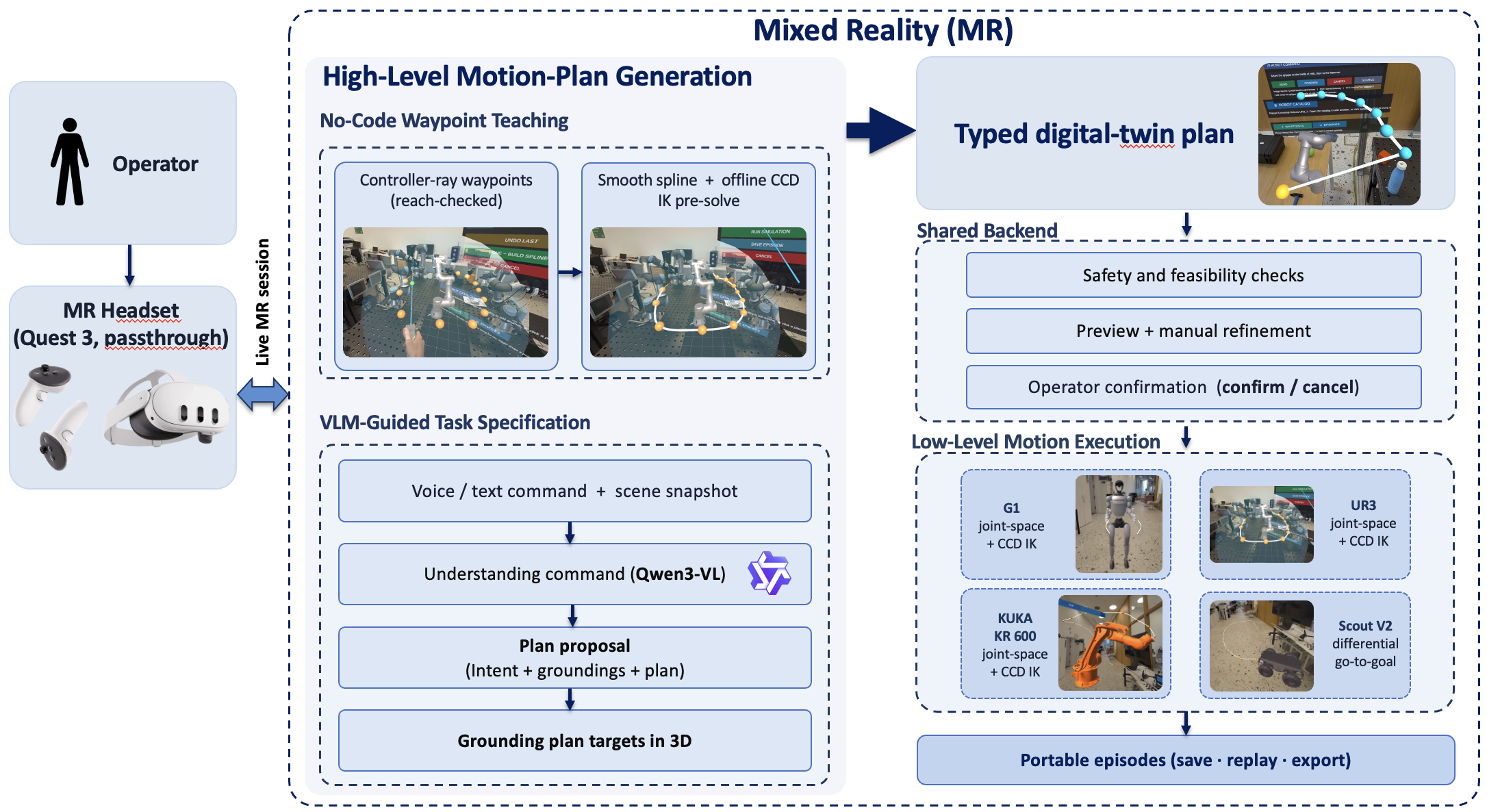}
  \caption{Overview of the ORCESTRA pipeline. The operator authors motions on co-located digital twins in MR through either no-code waypoint teaching or VLM-guided task specification (Qwen3-VL). Both modalities produce a typed digital-twin plan that passes through a shared backend -- safety checks, MR preview with manual refinement, and operator confirmation -- before being executed on UR3, KUKA KR 600, or AgileX Scout V2 and stored as a portable robot-relative episode.}
  \label{fig:pipeline}
  \vspace{-5mm}
\end{figure*}

\subsection{MR Digital-Twin Workspace}

ORCESTRA runs in desktop, immersive-VR, and MR passthrough modes; a mode manager switches camera, pointer, and passthrough configuration while world-space UI panels stay valid, so the same interaction logic tested on desktop deploys unchanged to the headset. The current implementation includes four embodiments:
Universal Robots UR3 and KUKA KR~600~FORTEC as 6-DOF fixed-base manipulators, AgileX Scout~V2 as a skid-steer mobile base, and Unitree G1 as a humanoid digital twin. 

Robot placement is performed directly in the MR scene: the operator selects a robot from the catalog, and ORCESTRA spawns a translucent ghost that follows the controller ray and lands on detected horizontal surfaces in MR mode (a virtual floor otherwise), with a fine-tuning panel for position and yaw before confirmation. The placed twin then exposes its own control panel and, for manipulators, a translucent reach visualization. 

\subsection{Robot Embodiments and Low-Level Control}

ORCESTRA separates task specification from embodiment-specific control: each placed robot exposes a common digital-twin abstraction to the no-code interface and the VLM-guided planner, while low-level execution is delegated to its own controller.

For 6-DOF manipulators, the implementation uses a joint-space controller and a cyclic-coordinate-descent (CCD) inverse-kinematics solver, supporting joint and TCP jogging and trajectory playback through joint drive targets. During replay, an offline IK pre-solve over sampled path points yields a joint trajectory, so
playback follows the previewed path without per-frame IK. The AgileX Scout~V2 mobile base uses a differential-drive go-to-goal controller with physics-based wheel contact: the operator or planner provides floor waypoints, and the controller drives the base between them.

\subsection{No-Code Waypoint Teaching and Episode Replay}

The no-code modality lets the operator program robot motion without writing code. Waypoints are created in the MR workspace with the controller ray; for manipulators, a 3D marker is pushed nearer or farther with the thumbstick to specify TCP targets, and targets outside the reachable workspace are rejected. For the mobile base, the floor becomes the teaching plane and the operator
specifies a sequence of route points. ORCESTRA then fits a smooth spline through the taught waypoints and performs an offline CCD IK pre-solve (a go-to-goal follower for the mobile base), so execution writes joint drive targets directly and the twin follows the previewed path.

Authored motions are stored as robot-relative JSON episodes:
\begin{equation}
E =
\Bigl(
r,\,
e,\,
T_0,\,
q_0,\,
\{p_i\}_{i=1}^{N},\,
\rho
\Bigr),
\label{eq:episode}
\end{equation}
where $r$ is the robot identifier, $e$ is the embodiment type, $T_{0}$ is the record-time base pose, $q_{0}$ is the recorded start configuration (joint angles for manipulators; empty for the mobile base), $\{p_i\}$ are the waypoints in the robot's record-time base frame, and $\rho$ is the playback rate. Because waypoints live in the base frame, an episode moves with the twin if repositioned; on load, the robot returns to~$q_{0}$ and replays relative to its current base pose. This also provides a basis for future export to a physical controller.

\subsection{VLM-Guided Task Specification}

The language-guided modality lets the operator describe a task by voice or text. The VLM acts as a high-level task interpreter and visual-reference module, not a direct controller: it returns a structured plan proposal and image-space references, while metric grounding, validation, preview, and execution remain on the client. On submission, ORCESTRA sends the headset view, a scene snapshot
(placed twins, camera, MR environment), the command, and optional audio through the FastAPI gateway to the VLM, which returns a typed JSON object:
\begin{equation}
\mathcal{R} =
\Bigl(
I,\,
\{g_j\}_{j=1}^{M},\,
P_{\mathrm{AI}},\,
D
\Bigr),
\label{eq:vlm_response}
\end{equation}
with parsed intent $I$, visual groundings $g_j$ (each a label, confidence $\kappa_j$, and image-space reference such as a bounding box or preferred point), an intermediate plan $P_{\mathrm{AI}}$ (kind, target robot, waypoints, and a contact-allowed flag $\alpha$), and diagnostics $D$. 

To convert an image-space reference $u$ into a metric 3D target, the client casts a ray $R_u(t) = o + t\,d(u)$ from the active camera and takes the first valid intersection with the MR scene:
\begin{equation}
x^{\star} =
\begin{cases}
R_u(t^{\star}), &
t^{\star} = \min \{t \mid R_u(t) \in \mathcal{C}\}, \\[2mm]
R_u(t_{\Pi}), &
\text{if no collider intersection exists and } R_u(t_{\Pi}) \in \Pi,
\end{cases}
\label{eq:grounding}
\end{equation}
where $\mathcal{C}$ denotes scene colliders (robot geometry, physics objects, MR planes) and $\Pi$ is the fallback floor plane; known labeled objects can also be grounded from stored world positions. The grounded targets become an embodiment-specific plan rendered as an editable preview, so the operator can adjust the proposed waypoints before confirming.

\subsection{Shared Grounding, Validation, Preview, and Execution}

Both modalities converge at the digital-twin validation and execution layer. Manual plans come directly from controller input; VLM-guided plans are first grounded into metric 3D targets. In either case, the motion is checked against the selected embodiment before the twin moves. For an AI-generated plan, we summarize the acceptance condition as
\begin{equation}
\begin{aligned}
\textsc{valid}(P_{\mathrm{AI}}) =
&\ \textsc{placed}(r)
\wedge \textsc{kind}(k,e)
\wedge \textsc{grounded}_{\tau}(\{g_j\})
\\
&\wedge\ \textsc{finite}(\{w_j\})
\wedge \textsc{feasible}_{e}(P_{\mathrm{AI}})
\wedge \neg\,\alpha,
\end{aligned}
\label{eq:validation}
\end{equation}
requiring that the selected robot is instantiated, the plan kind suits its embodiment, every visual reference is grounded above confidence $\tau$, all waypoint coordinates are finite, and the plan is realizable ($\textsc{feasible}_{e}$: reach plus IK pre-solve for manipulators, route validity with obstacle checks for the mobile base). Only valid plans are previewed; the operator inspects the targets or path and confirms or cancels, and execution starts only after confirmation. This confirmation-gated design is central to ORCESTRA: VLM output is a proposal to be grounded and verified on a co-located twin, not a direct command to the physical robot.

\section{Technical Validation}

We evaluated ORCESTRA as a complete, integrated MR prototype on Quest~3 with the VLM gateway on a workstation GPU. The goal is not to compare against teach-pendant programming, but to verify that the system works end-to-end across heterogeneous embodiments and both modalities.

\subsection{No-Code Teaching}

The no-code modality lets the operator program robot motion without writing
code. Waypoints are created in the MR workspace with the controller ray; for
manipulators, a 3D marker is pushed with the thumbstick to specify TCP targets,
and targets outside the reachable workspace are rejected. For the mobile base,
the floor becomes the teaching plane for a sequence of route points.

We tested this workflow on UR3, KUKA KR~600~FORTEC, and AgileX Scout~V2. For the
manipulators, the system supports placement, waypoint creation with reach
feedback, spline generation, offline IK pre-solving, preview, saving, loading,
and replay from a robot-relative episode; Scout~V2 uses the same interface to
produce a floor-plane route executed by the mobile-base controller. One
authoring pipeline thus supports both arm trajectories and mobile-base routes
without changing the interaction model.

\subsection{Language-Guided Grounding: User Validation}

We evaluated grounding accuracy in a preliminary user study. Since the prototype
has no gripper, this isolates VLM-side understanding rather than physical grasp
execution: whether the VLM correctly identifies the object referred to by the
command and proposes a visually plausible grasp point, from which the client
constructs and previews an approach trajectory on the twin. Five participants
each issued a relocation command for one object from each of four categories of
increasing complexity – simple geometric (e.g., a cube or a ball), everyday
(a water bottle), unconventional (a hand tool), and large, voluminous (a box)
– for 20 trials total. A trial was successful if (i)~the VLM identified the
correct object, (ii)~the proposed grasp point was visually plausible, and
(iii)~the client executed the corresponding approach trajectory on the twin.

\begin{table}[htb]
  \caption{Language-guided grounding success across object categories
  (5~participants, one trial per category).}
  \label{tab:userstudy}
  \scriptsize
  \centering
  \begin{tabular}{@{}l l c@{}}
  \toprule
  Object category & Example & Success \\
  \midrule
  Simple geometric    & cube, ball   & 5/5 \\
  Everyday            & water bottle & 5/5 \\
  Unconventional      & hand tool    & 4/5 \\
  Large / voluminous  & box          & 1/5 \\
  \bottomrule
  \end{tabular}
\end{table}

As summarized in Table~\ref{tab:userstudy}, simple geometric and everyday objects
were grounded and approached reliably across all participants. Performance
degraded moderately for unconventional objects, where occasional grasp-point or
recognition errors reduced the success rate. Large, voluminous objects proved
hardest: their size and geometry make a single, visually plausible grasp point
harder to define and approach, consistent with such objects typically requiring
bimanual or multi-contact rather than single-end-effector manipulation.

\begin{figure}[!t]
 \centering
 \begin{minipage}{0.49\columnwidth}
   \centering
   \includegraphics[width=\linewidth]{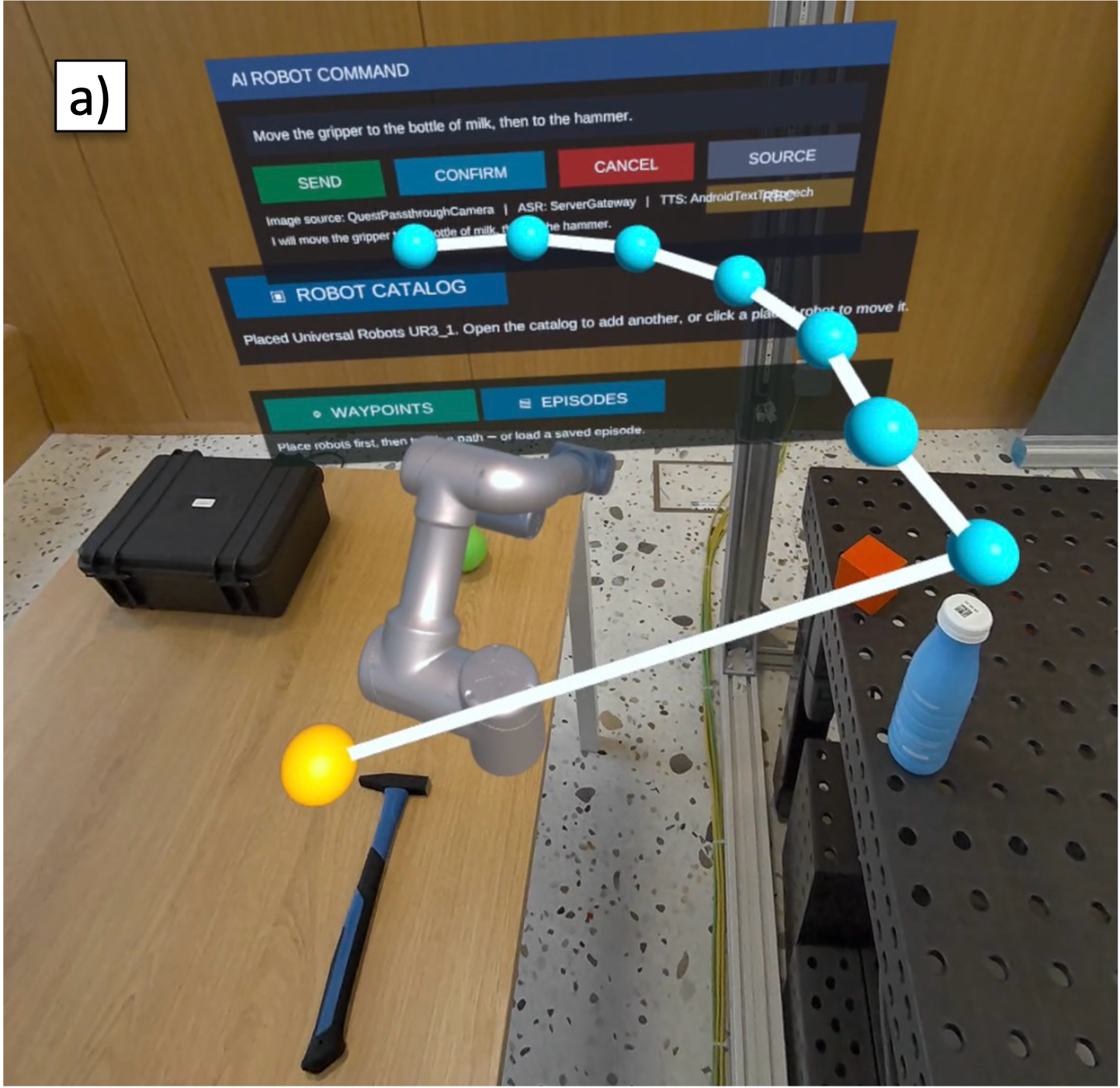}
 \end{minipage}
 \hfill
 \begin{minipage}{0.49\columnwidth}
   \centering
   \includegraphics[width=\linewidth]{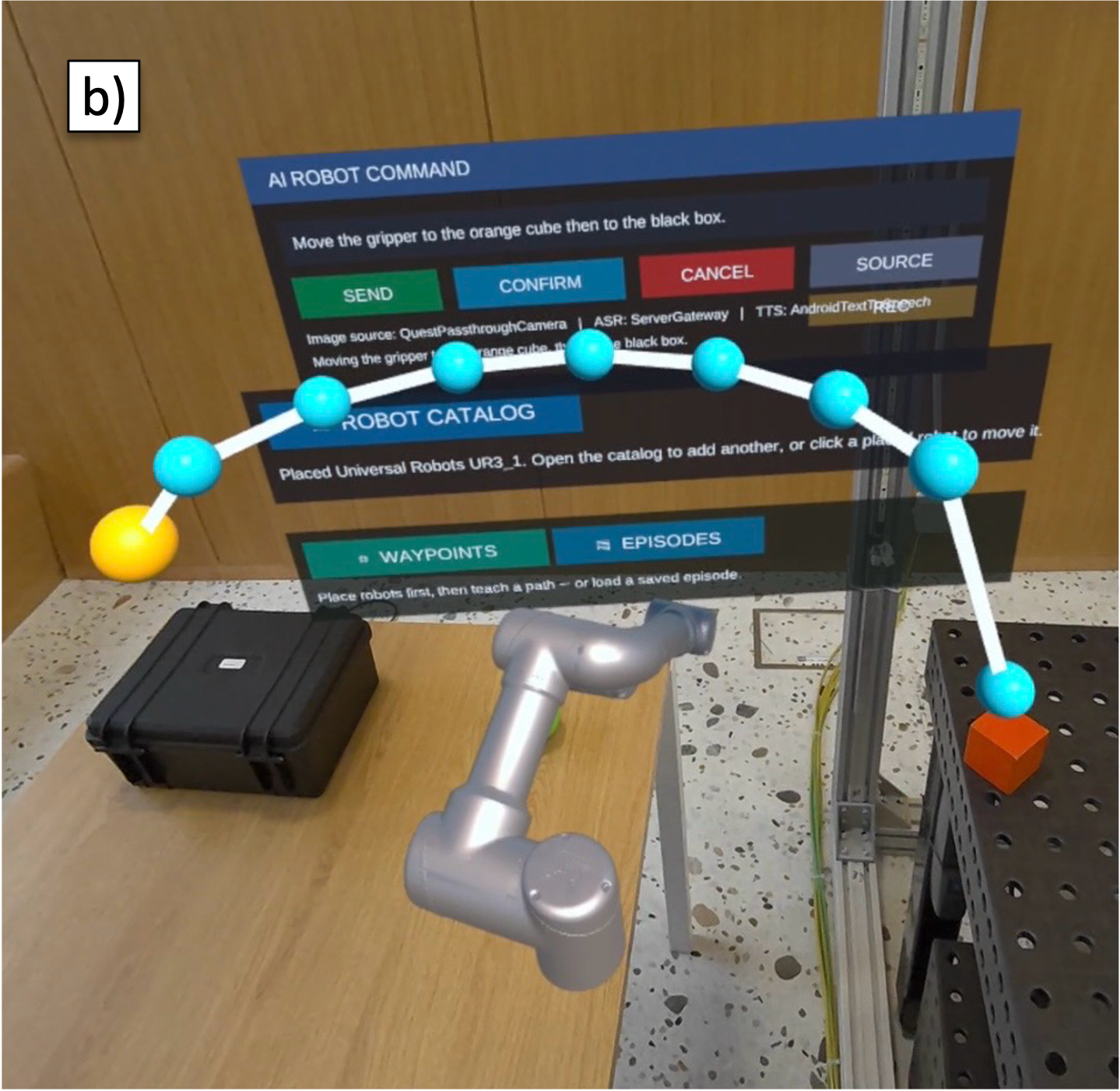}
 \end{minipage}
 \caption{Two language-guided digital-twin runs.
 (a) ``move the gripper to the bottle of milk, then to the hammer'';
 (b) ``move the gripper to the orange cube, then to the black box.''
 In each case the VLM grounds the referenced objects in the live view, the
 client lifts them to metric 3D and previews the approach trajectory through
 both targets (waypoints along the spline), and the twin executes only after
 operator confirmation.}
 \vspace{-5mm}
 \label{fig:results}
\end{figure}

Overall, the grounding pipeline is sufficiently reliable for common object
categories to serve as a practical input channel for digital-twin programming,
while exposing a clear boundary – large, geometrically complex objects –
where single-point grounding must be extended. Importantly, all failures were
caught at the preview stage, validating the confirmation-gated safety layer.

\section{Conclusion and Future Work}

We introduced ORCESTRA, an integrated MR system for programming robot digital
twins through no-code waypoint teaching and language-guided VLM control.
Running on Quest~3 across UR3, KUKA KR~600~FORTEC, AgileX Scout~V2, and Unitree
G1, it unifies both modalities behind one backend for grounding, validation,
preview, and confirmation-gated execution. A single plan representation carries
across robot types and input channels, while decoupling VLM reasoning from
metric execution renders the language module interchangeable and neutralizes
hallucinated outputs before any motion is committed.


Future work advances along three fronts. First, passthrough depth and scene meshes will supersede the horizontal-plane fallback. Second, constraint-based planning will let the language modality articulate finer spatial relations, while closing the loop with grippers of diverse morphologies – parallel-jaw,
multi-fingered, and vacuum will broaden the range of tractable objects and unlock bimanual tasks. Third, we will export episodes to physical controllers and stage a user study against teach-pendant programming.

\section*{Acknowledgements} 
Research reported in this publication was financially supported by the RSF grant No. 24-41-02039.


\bibliographystyle{abbrv-doi}
\bibliography{orcestra}

@article{qwen3vl,
  author  = {Bai, Shuai and Cai, Yuxuan and Chen, Ruizhe and Chen, Keqin and Chen, Xionghui and Cheng, Zesen and Deng, Lianghao and Ding, Wei and Gao, Chang and Ge, Chunjiang and others},
  title   = {{Q}wen3-{VL} Technical Report},
  journal = {arXiv preprint arXiv:2511.21631},
  year    = {2025}
}

@inproceedings{rekep,
  author    = {Huang, Wenlong and Wang, Chen and Li, Yunzhu and Zhang, Ruohan and Fei-Fei, Li},
  title     = {{R}e{K}ep: Spatio-Temporal Reasoning of Relational Keypoint Constraints for Robotic Manipulation},
  booktitle = {Proc. Conf. on Robot Learning (CoRL)},
  series    = {PMLR},
  volume    = {270},
  pages     = {4573--4602},
  year      = {2024}
}

@inproceedings{voxposer,
  author    = {Huang, Wenlong and Wang, Chen and Zhang, Ruohan and Li, Yunzhu and Wu, Jiajun and Li, Fei-Fei},
  title     = {{V}ox{P}oser: Composable {3D} Value Maps for Robotic Manipulation with Language Models},
  booktitle = {Conf. on Robot Learning (CoRL)},
  year      = {2023}
}

@inproceedings{ostanin2020mr,
  author    = {Ostanin, Mikhail and Mikhel, Stanislav and Evlampiev, Alexey and Skvortsova, Valeria and Klimchik, Alexandr},
  title     = {Human-Robot Interaction for Robotic Manipulator Programming in Mixed Reality},
  booktitle = {Proc. IEEE Int. Conf. on Robotics and Automation (ICRA)},
  pages     = {2805--2811},
  year      = {2020},
  doi     = {10.1109/ICRA40945.2020.9196965}
}

@article{mr_cobot_dt,
  author  = {Garg, Gaurav and Kuts, Vladimir and Anbarjafari, Gholamreza},
  title   = {Digital Twin for {FANUC} Robots: Industrial Robot Programming and Simulation Using Virtual Reality},
  journal = {Sustainability},
  volume  = {13},
  number  = {18},
  pages   = {10336},
  year    = {2021},
  doi     = {10.3390/su131810336}
}

@inproceedings{warevr,
  author    = {Kalinov, Ivan and Trinitatova, Daria and Tsetserukou, Dzmitry},
  title     = {{W}are{VR}: Virtual Reality Interface for Supervision of Autonomous Robotic System Aimed at Warehouse Stocktaking},
  booktitle = {Proc. IEEE Int. Conf. on Systems, Man, and Cybernetics (SMC)},
  pages     = {2139--2145},
  year      = {2021},
  doi       = {10.1109/SMC52423.2021.9659133}
}

@inproceedings{humanoidvlm,
  author    = {Mahmoud, Yara and Yaqoot, Yasheerah and Altamirano Cabrera, Miguel and Tsetserukou, Dzmitry},
  title     = {{H}umanoid{VLM}: Vision--Language--Guided Impedance Control for Contact-Rich Humanoid Manipulation},
  booktitle = {Proc. ACM/IEEE Int. Conf. on Human-Robot Interaction (HRI)},
  pages     = {918--922},
  year      = {2026}
}

@inproceedings{suzuki2022arsurvey,
  author    = {Suzuki, Ryo and Karim, Adnan and Xia, Tian and Hedayati, Hooman and Marquardt, Nicolai},
  title     = {Augmented Reality and Robotics: A Survey and Taxonomy for {AR}-Enhanced Human-Robot Interaction and Robotic Interfaces},
  booktitle = {Proc. CHI Conf. on Human Factors in Computing Systems (CHI)},
  pages     = {1--33},
  year      = {2022},
  doi       = {10.1145/3491102.3517719}
}

@inproceedings{quintero2018augtraj,
  author    = {Quintero, Camilo Perez and Li, Sarah and Pan, Matthew K. X. J. and Chan, Wesley P. and Van der Loos, H. F. Machiel and Croft, Elizabeth},
  title     = {Robot Programming Through Augmented Trajectories in Augmented Reality},
  booktitle = {Proc. IEEE/RSJ Int. Conf. on Intelligent Robots and Systems (IROS)},
  pages     = {1838--1844},
  year      = {2018},
  doi       = {10.1109/IROS.2018.8593700}
}

@article{fang2024waypoint,
  author  = {Fang, Cathy Mengying and Zieli\'nski, Krzysztof and Maes, Pattie and Paradiso, Joe and Blumberg, Bruce and Kj{\ae}rgaard, Mikkel Baun},
  title   = {Enabling Waypoint Generation for Collaborative Robots Using {LLM}s and Mixed Reality},
  journal = {arXiv preprint arXiv:2403.09308},
  year    = {2024}
}

\end{document}